\documentstyle[aclap]{article}

\title{\vspace{-0.5in}
Exemplar-Based Word Sense Disambiguation: \\
Some Recent Improvements}

\author{
Hwee Tou Ng \\
DSO National Laboratories \\
20 Science Park Drive \\
Singapore 118230 \\ 
{\tt nhweetou@dso.org.sg}
}

\begin{document}
\bibliographystyle{fullname}
\maketitle
\vspace{-0.5in}
\begin{abstract}

In this paper, we report recent improvements to the exemplar-based
learning approach for word sense disambiguation that have achieved
higher disambiguation accuracy.  By using a larger value of $k$, the
number of nearest neighbors to use for determining the class of a test
example, and through 10-fold cross validation to automatically
determine the best $k$, we have obtained improved disambiguation
accuracy on a large sense-tagged corpus first used in \cite{ng96}.
The accuracy achieved by our improved exemplar-based classifier is
comparable to the accuracy on the same data set obtained by the
Naive-Bayes algorithm, which was reported in \cite{mooney96} to have
the highest disambiguation accuracy among seven state-of-the-art
machine learning algorithms.

\end{abstract}

\section{Introduction}

Much recent research on word sense disambiguation (WSD) has adopted a
corpus-based, learning approach. Many different learning approaches
have been used, including neural networks \cite{leacock93},
probabilistic algorithms
\cite{bruce94,gale92a,gale95,leacock93,yarowsky92}, decision lists
\cite{yarowsky94}, exemplar-based learning algorithms
\cite{cardie93,ng96}, etc.

In particular, Mooney \shortcite{mooney96} evaluated seven
state-of-the-art machine learning algorithms on a common data set for
disambiguating six senses of the word ``line''. The seven algorithms
that he evaluated are: a Naive-Bayes classifier \cite{duda73}, a
perceptron \cite{rosenblatt58}, a decision-tree learner
\cite{quinlan93}, a k nearest-neighbor classifier (exemplar-based
learner) \cite{cover67}, logic-based DNF and CNF learners
\cite{mooney95}, and a decision-list learner \cite{rivest87}. His
results indicate that the simple Naive-Bayes algorithm gives the
highest accuracy on the ``line'' corpus tested. Past research in
machine learning has also reported that the Naive-Bayes algorithm
achieved good performance on other machine learning tasks
\cite{clark89,kohavi96}. This is in spite of the conditional
independence assumption made by the Naive-Bayes algorithm, which may
be unjustified in the domains tested.  Gale, Church and Yarowsky
\cite{gale92a,gale95,yarowsky92} have also successfully
used the Naive-Bayes algorithm (and several extensions and variations)
for word sense disambiguation.

On the other hand, our past work on WSD \cite{ng96} used an
exemplar-based (or nearest neighbor) learning approach. Our WSD
program, {\sc Lexas}, extracts a set of features, including part of
speech and morphological form, surrounding words, local collocations,
and verb-object syntactic relation from a sentence containing the word
to be disambiguated. These features from a sentence form an
example. {\sc Lexas} then uses the exemplar-based learning algorithm
{\sc Pebls} \cite{cost93} to find the sense (class) of the word to be
disambiguated.

In this paper, we report recent improvements to the exemplar-based
learning approach for WSD that have achieved higher disambiguation
accuracy.  The exemplar-based learning algorithm {\sc Pebls} contains
a number of parameters that must be set before running the
algorithm. These parameters include the number of nearest neighbors to
use for determining the class of a test example (i.e., $k$ in a $k$
nearest-neighbor classifier), exemplar weights, feature weights, etc.
We found that the number $k$ of nearest neighbors used has a
considerable impact on the accuracy of the induced exemplar-based
classifier. By using 10-fold cross validation \cite{kohavi95} on the
training set to automatically determine the best $k$ to use, we have
obtained improved disambiguation accuracy on a large sense-tagged
corpus first used in \cite{ng96}. The accuracy achieved by our
improved exemplar-based classifier is comparable to the accuracy on
the same data set obtained by the Naive-Bayes algorithm, which was
reported in \cite{mooney96} to have the highest disambiguation
accuracy among seven state-of-the-art machine learning algorithms.

The rest of this paper is organized as follows.
Section~\ref{sect:algm} gives a brief description of the
exemplar-based algorithm {\sc Pebls} and the Naive-Bayes
algorithm. Section~\ref{sect:improve} describes the 10-fold cross
validation training procedure to determine the best $k$ number of
nearest neighbors to use. Section~\ref{sect:results} presents the
disambiguation accuracy of {\sc Pebls} and Naive-Bayes on the large
corpus of \cite{ng96}. Section~\ref{sect:discussion} discusses the
implications of the results. Section~\ref{sect:conclusion} gives the
conclusion.

\section{Learning Algorithms}
\label{sect:algm}

\subsection{\sc{Pebls}}

The heart of exemplar-based learning is a measure of the similarity,
or distance, between two examples. If the distance between two
examples is small, then the two examples are similar. In {\sc Pebls}
\cite{cost93}, the distance between two symbolic values $v_1$ and
$v_2$ of a feature $f$ is defined as:

\[ d(v_1, v_2) = \sum_{i=1}^{n} | P(C_i|v_1) - P(C_i|v_2) | \]

\noindent where $n$ is the total number of classes.
$P(C_i|v_1)$ is estimated by $\frac{N_{1,i}}{N_1}$, where $N_{1,i}$ is
the number of training examples with value $v_1$ for feature $f$ that
is classified as class $i$ in the training corpus, and $N_1$ is the
number of training examples with value $v_1$ for feature $f$ in any
class. $P(C_i|v_2)$ is estimated similarly. This distance metric of
{\sc Pebls} is adapted from the value difference metric of the earlier
work of \cite{stanfill86}.  The distance between two examples is the
sum of the distances between the values of all the features of the two
examples.

Let $k$ be the number of nearest neighbors to use for determining the
class of a test example, $k \geq 1$.  During testing, a test example
is compared against {\em all} the training examples.  {\sc Pebls} then
determines the $k$ training examples with the shortest distance to the
test example. Among these $k$ closest matching training examples, the
class which the majority of these $k$ examples belong to will be
assigned as the class of the test example, with tie among multiple
majority classes broken randomly.

Note that the nearest neighbor algorithm tested in \cite{mooney96}
uses Hamming distance as the distance metric between two symbolic
feature values. This is different from the above distance metric used
in {\sc Pebls}.

\subsection{Naive-Bayes}

Our presentation of the Naive-Bayes algorithm \cite{duda73} follows
that of \cite{clark89}. This algorithm is based on Bayes' theorem:

\[ P(C_i|\wedge v_j) = \frac{P(\wedge v_j|C_i)P(C_i)}{P(\wedge v_j)} \hspace{1.5cm} i = 1 \ldots n \]

\noindent where $P(C_i|\wedge v_j)$ is the probability that a test
example is of class $C_i$ given feature values $v_j$. ($\wedge v_j$
denotes the conjunction of all feature values in the test example.)
The goal of a Naive-Bayes classifier is to determine the class $C_i$
with the highest conditional probability $P(C_i|\wedge v_j)$. Since
the denominator $P(\wedge v_j)$ of the above expression is constant
for all classes $C_i$, the problem reduces to finding the class $C_i$
with the maximum value for the numerator.

The Naive-Bayes classifier assumes independence of example features,
so that

\[ P(\wedge v_j|C_i) = \prod_{j} P(v_j|C_i) \]

During training, Naive-Bayes constructs the matrix $P(v_j|C_i)$, and
$P(C_i)$ is estimated from the distribution of training examples among
the classes. To avoid one zero count of $P(v_j|C_i)$ nullifying the
effect of the other non-zero conditional probabilities in the
multiplication, we replace zero counts of $P(v_j|C_i)$ by $P(C_i)/N$,
where $N$ is the total number of training examples. Other more complex
smoothing procedures (such as those used in \cite{gale92a}) are also
possible, although we have not experimented with these other
variations.

For the experimental results reported in this paper, we used the
implementation of Naive-Bayes algorithm in the {\sc Pebls} program
\cite{rachlin93}, which has an option for training and testing using
the Naive-Bayes algorithm. We only changed the handling of zero
probability counts to the method just described.

\section{Improvements to Exemplar-Based WSD}
\label{sect:improve}

{\sc Pebls} contains a number of parameters that must be set before
running the algorithm. These parameters include $k$ (the number of
nearest neighbors to use for determining the class of a test example),
exemplar weights, feature weights, etc.  Each of these parameters has
a default value in {\sc Pebls}, eg., $k = 1$, no exemplar weighting,
no feature weighting, etc. We have used the default values for all
parameter settings in our previous work on exemplar-based WSD reported
in \cite{ng96}. However, our preliminary investigation indicates that,
among the various learning parameters of {\sc Pebls}, the number $k$
of nearest neighbors used has a considerable impact on the accuracy of
the induced exemplar-based classifier.

Cross validation is a well-known technique that can be used for
estimating the expected error rate of a classifier which has been
trained on a particular data set. For instance, the C4.5 program
\cite{quinlan93} contains an option for running cross validation to
estimate the expected error rate of an induced rule set.  Cross
validation has been proposed as a general technique to automatically
determine the parameter settings of a given learning algorithm using a
particular data set as training data \cite{kohavi95}.

In $m$-fold cross validation, a training data set is partitioned into
$m$ (approximately) equal-sized blocks, and the learning algorithm is
run $m$ times. In each run, one of the $m$ blocks of training data is
set aside as test data (the holdout set) and the remaining $m - 1$
blocks are used as training data. The average error rate of the $m$
runs is a good estimate of the error rate of the induced classifier.

For a particular parameter setting, we can run $m$-fold cross
validation to determine the expected error rate of that particular
parameter setting. We can then choose an optimal parameter setting
that minimizes the expected error rate. Kohavi and John
\shortcite{kohavi95} reported the effectiveness of such a technique in
obtaining optimal sets of parameter settings over a large number of
machine learning problems.

In our present study, we used 10-fold cross validation to
automatically determine the best $k$ (number of nearest neighbors) to
use from the training data. To determine the best $k$ for
disambiguating a word on a particular training set, we run 10-fold
cross validation using {\sc Pebls} 21 times, each time with $k = 1, 5,
10, 15, \ldots, 85, 90, 95, 100$. We compute the error rate for each
$k$, and choose the value of $k$ with the minimum error rate. Note
that the automatic determination of the best $k$ through 10-fold cross
validation makes use of {\em only} the training set, without looking
at the test set at all.

\section{Experimental Results}
\label{sect:results}

Mooney \shortcite{mooney96} has reported that the Naive-Bayes
algorithm gives the best performance on disambiguating six senses of
the word ``line'', among seven state-of-the-art learning algorithms
tested. However, his comparative study is done on only one word using
a data set of 2,094 examples.  In our present study, we evaluated {\sc
Pebls} and Naive-Bayes on a much larger corpus containing sense-tagged
occurrences of 121 nouns and 70 verbs. This corpus was first reported
in \cite{ng96}, and it contains about 192,800 sense-tagged word
occurrences of 191 most frequently occurring and ambiguous words of
English.\footnote{This corpus is available from the Linguistic Data
Consortium (LDC). Contact the LDC at ldc@unagi.cis.upenn.edu for
details.} These 191 words have been tagged with senses from {\sc
WordNet} \cite{miller90}, an on-line, electronic dictionary available
publicly. For this set of 191 words, the average number of senses per
noun is 7.8, while the average number of senses per verb is 12.0.  The
sentences in this corpus were drawn from the combined corpus of the 1
million word Brown corpus and the 2.5 million word Wall Street Journal
(WSJ) corpus.

We tested both algorithms on two test sets from this corpus. The first
test set, named BC50, consists of 7,119 occurrences of the 191 words
appearing in 50 text files of the Brown corpus. The second test set,
named WSJ6, consists of 14,139 occurrences of the 191 words appearing
in 6 text files of the WSJ corpus. Both test sets are identical to the
ones reported in \cite{ng96}.

Since the primary aim of our present study is the comparative
evaluation of learning algorithms, not feature representation, we have
chosen, for simplicity, to use local collocations as the only features
in the example representation. Local collocations have been found to
be the single most informative set of features for WSD \cite{ng96}.
That local collocation knowledge provides important clues to WSD has
also been pointed out previously by Yarowsky \shortcite{yarowsky93}.

Let $w$ be the word to be disambiguated, and let $l_2$ $l_1$ $w$ $r_1$
$r_2$ be the sentence fragment containing $w$. In the present study,
we used seven features in the representation of an example, which are
the local collocations of the surrounding 4 words. These seven features
are: $l_2$\_$l_1$, $l_1$\_$r_1$, $r_1$\_$r_2$, $l_1$, $r_1$, $l_2$,
and $r_2$. The first three features are concatenation of two
words.\footnote{The first five of these seven features were also used
in \cite{ng96}.}

\begin{table}
\centering
\begin{tabular}{|l|c|c|} \hline
\multicolumn{1}{|c|}{Algorithm} & BC50   & WSJ6   \\ \hline
Sense 1                         & 40.5\% & 44.8\% \\ \hline
Most Frequent                   & 47.1\% & 63.7\% \\ \hline
Ng \& Lee (1996)                & 54.0\% & 68.6\% \\ \hline
{\sc Pebls} ($k = 1$)           & 55.0\% & 70.2\% \\ \hline
{\sc Pebls} ($k = 20$)          & 58.5\% & 74.5\% \\ \hline
{\sc Pebls} (10-fold c.v.)      & 58.7\% & 75.2\% \\ \hline
Naive-Bayes                     & 58.2\% & 74.5\% \\ \hline
\end{tabular}
\caption{Experimental Results}
\label{tab:results}
\end{table}

The experimental results obtained are tabulated in
Table~\ref{tab:results}. The first three rows of accuracy figures are
those of \cite{ng96}. The default strategy of picking the most
frequent sense has been advocated as the baseline performance for
evaluating WSD programs \cite{gale92b,miller94}. There are two
instantiations of this strategy in our current evaluation. Since {\sc
WordNet} orders its senses such that sense 1 is the most frequent
sense, one possibility is to always pick sense 1 as the best sense
assignment. This assignment method does not even need to look at the
training examples. We call this method ``Sense 1'' in
Table~\ref{tab:results}. Another assignment method is to determine the
most frequently occurring sense in the training examples, and to
assign this sense to all test examples.  We call this method ``Most
Frequent'' in Table~\ref{tab:results}.

The accuracy figures of {\sc Lexas} as reported in \cite{ng96} are
reproduced in the third row of Table~\ref{tab:results}. These figures
were obtained using all features including part of speech and
morphological form, surrounding words, local collocations, and
verb-object syntactic relation. However, the feature value pruning
method of \cite{ng96} only selects surrounding words and local
collocations as feature values if they are indicative of some sense
class as measured by conditional probability (See \cite{ng96} for
details).

The next three rows show the accuracy figures of {\sc Pebls} using the
parameter setting of $k = 1$, $k = 20$, and 10-fold cross validation
for finding the best $k$, respectively. The last row shows the
accuracy figures of the Naive-Bayes algorithm. Accuracy figures of the
last four rows are all based on only seven collocation features as
described earlier in this section. However, all possible feature
values (collocated words) are used, without employing the feature
value pruning method used in \cite{ng96}.

Note that the accuracy figures of {\sc Pebls} with $k = 1$ are 1.0\%
and 1.6\% higher than the accuracy figures of \cite{ng96} in the third
row, also with $k = 1$. The feature value pruning method of
\cite{ng96} is intended to keep only feature values deemed important
for classification. It seems that the pruning method has filtered out
some useful collocation values that improve classification accuracy,
such that this unfavorable effect outweighs the additional set of
features (part of speech and morphological form, surrounding words,
and verb-object syntactic relation) used.

Our results indicate that although Naive-Bayes performs better than
{\sc Pebls} with $k = 1$, {\sc Pebls} with $k = 20$ achieves
comparable performance. Furthermore, {\sc Pebls} with 10-fold cross
validation to select the best $k$ yields results slightly better than
the Naive-Bayes algorithm.

\section{Discussion}
\label{sect:discussion}

To understand why larger values of $k$ are needed, we examined the
performance of {\sc Pebls} when tested on the WSJ6 test set. During
10-fold cross validation runs on the training set, for each of the 191
words, we compared two error rates: the minimum expected error rate of
{\sc Pebls} using the best $k$, and the expected error rate of the
most frequent classifier. We found that for 13 words out of the 191
words, the minimum expected error rate of {\sc Pebls} using the best
$k$ is still higher than the expected error rate of the most frequent
classifier. That is, for these 13 words, {\sc Pebls} will produce, on
average, lower accuracy than the most frequent classifier.

Importantly, for 11 of these 13 words, the best $k$ found by {\sc
Pebls} are at least 85 and above. This indicates that for a training
data set when {\sc Pebls} has trouble even outperforming the most
frequent classifier, it will tend to use a large value for $k$. This
is explainable since for a large value of $k$, {\sc Pebls} will tend
towards the performance of the most frequent classifier, as it will
find the $k$ closest matching training examples and select the
majority class among this large number of $k$ examples. Note that in
the extreme case when $k$ equals the size of the training set, {\sc
Pebls} will behave exactly like the most frequent classifier.

Our results indicate that although {\sc Pebls} with $k = 1$ gives
lower accuracy compared with Naive-Bayes, {\sc Pebls} with $k = 20$
performs as well as Naive-Bayes. Furthermore, {\sc Pebls} with
automatically selected $k$ using 10-fold cross validation gives
slightly higher performance compared with Naive-Bayes. We believe that
this result is significant, in light of the fact that Naive-Bayes has
been found to give the best performance for WSD among seven
state-of-the-art machine learning algorithms \cite{mooney96}. It
demonstrates that an exemplar-based learning approach is suitable for
the WSD task, achieving high disambiguation accuracy.

One potential drawback of an exemplar-based learning approach is the
testing time required, since each test example must be compared with
every training example, and hence the required testing time grows
linearly with the size of the training set. However, more
sophisticated indexing methods such as that reported in
\cite{friedman77} can reduce this to logarithmic expected time, which
will significantly reduce testing time.

In the present study, we have focused on the comparison of learning
algorithms, but not on feature representation of examples. Our past
work \cite{ng96} suggests that multiple sources of knowledge are indeed
useful for WSD. Future work will explore the addition of these other
features to further improve disambiguation accuracy.

Besides the parameter $k$, {\sc Pebls} also contains other learning
parameters such as exemplar weights and feature weights. Exemplar
weighting has been found to improve classification performance
\cite{cost93}. Also, given the relative importance of the various
knowledge sources as reported in \cite{ng96}, it may be possible to
improve disambiguation performance by introducing feature weighting.
Future work can explore the effect of exemplar weighting and feature
weighting on disambiguation accuracy.

\section{Conclusion}
\label{sect:conclusion}

In summary, we have presented improvements to the exemplar-based
learning approach for WSD. By using a larger value of $k$, the number
of nearest neighbors to use for determining the class of a test
example, and through 10-fold cross validation to automatically
determine the best $k$, we have obtained improved disambiguation
accuracy on a large sense-tagged corpus.  The accuracy achieved by our
improved exemplar-based classifier is comparable to the accuracy on
the same data set obtained by the Naive-Bayes algorithm, which was
recently reported to have the highest disambiguation accuracy among
seven state-of-the-art machine learning algorithms.

\section{Acknowledgements}

Thanks to Ray Mooney for helpful discussions, and the anonymous
reviewers for their comments.


\begin{thebibliography}{fullname}

\bibitem[\protect\citename{Bruce and Wiebe}1994]{bruce94}
Rebecca Bruce and Janyce Wiebe.
\newblock 1994.
\newblock Word-sense disambiguation using decomposable models.
\newblock In {\em Proceedings of the 32nd Annual Meeting of the Association 
  for Computational Linguistics}, Las Cruces, New Mexico.

\bibitem[\protect\citename{Cardie}1993]{cardie93}
Claire Cardie.
\newblock 1993.
\newblock A case-based approach to knowledge acquisition for 
domain-specific sentence analysis.
\newblock In {\em Proceedings of the Eleventh National Conference
  on Artificial Intelligence}, pages 798--803, Washington, DC.

\bibitem[\protect\citename{Clark and Niblett}1989]{clark89}
Peter Clark and Tim Niblett.
\newblock 1989.
\newblock The {CN2} induction algorithm.
\newblock {\em Machine Learning}, 3(4):261--283.

\bibitem[\protect\citename{Cost and Salzberg}1993]{cost93}
Scott Cost and Steven Salzberg.
\newblock 1993.
\newblock A weighted nearest neighbor algorithm for learning with 
symbolic features.
\newblock {\em Machine Learning}, 10(1):57--78.

\bibitem[\protect\citename{Cover and Hart}1967]{cover67}
T. M. Cover and P. Hart.
\newblock 1967.
\newblock Nearest neighbor pattern classification.
\newblock {\em IEEE Transactions on Information Theory}, 13(1):21--27.

\bibitem[\protect\citename{Duda and Hart}1973]{duda73}
Richard Duda and Peter Hart.
\newblock 1973.
\newblock {\em Pattern Classification and Scene Analysis}. Wiley, New York.

\bibitem[\protect\citename{Friedman \bgroup et al.\egroup }1977]{friedman77}
J. Friedman, J. Bentley, and R. Finkel.
\newblock 1977.
\newblock An algorithm for finding best matches in logarithmic
expected time.
\newblock {\em ACM Transactions on Mathematical Software}, 3(3):209--226.

\bibitem[\protect\citename{Gale \bgroup et al.\egroup }1992a]{gale92a}
William Gale, Kenneth Ward Church, and David Yarowsky.
\newblock 1992a.
\newblock A Method for Disambiguating Word Senses in a Large Corpus.
\newblock {\em Computers and the Humanities}, 26:415--439.

\bibitem[\protect\citename{Gale \bgroup et al.\egroup }1992b]{gale92b}
William Gale, Kenneth Ward Church, and David Yarowsky.
\newblock 1992b.
\newblock Estimating upper and lower bounds on the performance 
of word-sense disambiguation programs.
\newblock In {\em Proceedings of the 30th Annual Meeting of the Association 
  for Computational Linguistics}, Newark, Delaware.

\bibitem[\protect\citename{Gale \bgroup et al.\egroup }1995]{gale95}
William Gale, Kenneth Ward Church, and David Yarowsky.
\newblock 1995.
\newblock Discrimination Decisions for 100,000 Dimensional Spaces.
\newblock {\em Annals of Operations Research}, 55:323--344.

\bibitem[\protect\citename{Kohavi and John}1995]{kohavi95}
Ron Kohavi and George~H. John.
\newblock 1995.
\newblock Automatic parameter selection by minimizing estimated error.
\newblock In {\em Machine Learning: Proceedings of the Twelfth
International Conference}.

\bibitem[\protect\citename{Kohavi}1996]{kohavi96}
Ron Kohavi.
\newblock 1996.
\newblock Scaling up the accuracy of {Naive-Bayes} classifiers: A
decision-tree hybrid.
\newblock In {\em Proceedings of the Second International Conference
on Knowledge Discovery and Data Mining}.

\bibitem[\protect\citename{Leacock \bgroup et al.\egroup }1993]{leacock93}
Claudia Leacock, Geoffrey Towell, and Ellen Voorhees.
\newblock 1993.
\newblock Corpus-based statistical sense resolution.
\newblock In {\em Proceedings of the {ARPA} Human Language Technology 
Workshop}.

\bibitem[\protect\citename{Miller}1990]{miller90}
George~A. Miller, Ed.
\newblock 1990.
\newblock {WordNet:} An on-line lexical database.
\newblock {\em International Journal of Lexicography}, 3(4):235--312.

\bibitem[\protect\citename{Miller \bgroup et al.\egroup }1994]{miller94}
George~A. Miller, Martin Chodorow, Shari Landes, Claudia Leacock, and 
Robert~G. Thomas.
\newblock 1994.
\newblock Using a semantic concordance for sense identification.
\newblock In {\em Proceedings of the {ARPA} Human Language Technology 
Workshop}.

\bibitem[\protect\citename{Mooney}1995]{mooney95}
Raymond~J. Mooney.
\newblock 1995.
\newblock Encouraging experimental results on learning CNF.
\newblock {\em Machine Learning}, 19(1):79--92.

\bibitem[\protect\citename{Mooney}1996]{mooney96}
Raymond~J. Mooney.
\newblock 1996.
\newblock Comparative experiments on disambiguating word senses: An
illustration of the role of bias in machine learning.
\newblock In {\em Proceedings of the Conference on Empirical
Methods in Natural Language Processing (EMNLP)}.

\bibitem[\protect\citename{Ng and Lee}1996]{ng96}
Hwee Tou Ng and Hian Beng Lee.
\newblock 1996.
\newblock Integrating multiple knowledge sources to disambiguate word
sense: An exemplar-based approach.
\newblock In {\em Proceedings of the 34th Annual Meeting of the
Association for Computational Linguistics (ACL)}, pages 40--47.

\bibitem[\protect\citename{Quinlan}1993]{quinlan93}
J. Ross Quinlan.
\newblock 1993.
\newblock {\em {C4.5:} Programs for Machine Learning}. Morgan
Kaufmann, San Mateo, CA.

\bibitem[\protect\citename{Rachlin and Salzberg}1993]{rachlin93}
John Rachlin and Steven Salzberg.
\newblock 1993.
\newblock {PEBLS} 3.0 User's Guide.

\bibitem[\protect\citename{Rivest}1987]{rivest87}
R. L. Rivest.
\newblock 1987.
\newblock Learning decision lists.
\newblock {\em Machine Learning}, 2(3):229--246.

\bibitem[\protect\citename{Rosenblatt}1958]{rosenblatt58}
F. Rosenblatt.
\newblock 1958.
\newblock The perceptron: A probabilistic model for information storage
and organization in the brain.
\newblock {\em Psychological Review}, 65:386--408.

\bibitem[\protect\citename{Stanfill and Waltz}1986]{stanfill86}
C Stanfill and David Waltz.
\newblock 1986.
\newblock Toward memory-based reasoning.
\newblock {\em Communications of the {ACM}}, 29(12):1213--1228.

\bibitem[\protect\citename{Yarowsky}1992]{yarowsky92}
David Yarowsky.
\newblock 1992.
\newblock Word-sense disambiguation using statistical models of 
{Roget's} categories trained on large corpora.
\newblock In {\em Proceedings of the Fifteenth International Conference 
on Computational Linguistics}, pages 454--460, Nantes, France.

\bibitem[\protect\citename{Yarowsky}1993]{yarowsky93}
David Yarowsky.
\newblock 1993.
\newblock One sense per collocation.
\newblock In {\em Proceedings of the {ARPA} Human Language Technology 
Workshop}.

\bibitem[\protect\citename{Yarowsky}1994]{yarowsky94}
David Yarowsky.
\newblock 1994.
\newblock Decision lists for lexical ambiguity resolution: Application
to accent restoration in {Spanish} and {French}.
\newblock In {\em Proceedings of the 32nd Annual Meeting of the Association 
  for Computational Linguistics}, Las Cruces, New Mexico.

\end{thebibliography}
\end{document}